\documentclass[10pt,twocolumn,letterpaper]{article}

\usepackage{cvpr}              
\definecolor{cvprblue}{rgb}{0.21,0.49,0.74}
\usepackage[pagebackref,breaklinks,colorlinks,allcolors=cvprblue]{hyperref}

\usepackage{etoc}
\usepackage{etoc} 
\usepackage{titletoc}
\usepackage[subfigure]{tocloft} 

\usepackage{algorithm}
\usepackage{algpseudocode}  
\usepackage{float}          

\usepackage{epsfig}
\usepackage{graphicx}
\usepackage{amsmath}
\usepackage{amssymb}
\usepackage{booktabs}
\usepackage{multirow}
\usepackage{tabularx}
\usepackage{enumitem}
\usepackage{gensymb}
\usepackage{colortbl}
\usepackage{caption}
\usepackage{makecell}
\usepackage{wrapfig}
\usepackage{amssymb}
\usepackage{pifont}
\definecolor{mycyan}{cmyk}{.1,0,0,0}
\newcommand{\mypara}[1]{\vspace{1mm}\noindent\textbf{#1}}
\usepackage[capitalize]{cleveref}
\definecolor{mygray}{gray}{.95}
\definecolor{myblue}{HTML}{ECF2F8}
\definecolor{mycyan}{cmyk}{.1,0,0,0}
\definecolor{mygray}{gray}{.95}
\definecolor{mypink}{rgb}{.99,.91,.95}
\definecolor{mypurple}{rgb}{0,0,0}

\newcommand{\name}{AtomicVLA}
\newcommand{\tb}[1]{\textbf{#1}}

\def\paperID{} 
\def\confName{CVPR}
\def\confYear{2026}

\title{AtomicVLA: Unlocking the Potential of Atomic Skill Learning in Robots}

\author{
  Likui Zhang$^{1}$ \quad
  Tao Tang$^{1}$ \quad
  Zhihao Zhan$^{1}$ \quad
  Xiuwei Chen$^{1}$ \quad 
  Zisheng Chen$^{1}$ \quad 
  Jianhua Han$^{3}$ \quad \\
  Jiangtong Zhu$^{3}$ \quad 
  Pei Xu$^{3}$ \quad
  Hang Xu$^{3}$ \quad
  Hefeng Wu$^{1}$ \quad
  Liang Lin$^{1}$ \footnotemark[2]\quad
  Xiaodan Liang$^{1,2}$ \footnotemark[2] \quad
  \\
\fontsize{11pt}{\baselineskip}\selectfont
$^1$Sun Yat-sen University   
\fontsize{11pt}{\baselineskip}\selectfont
$^2$Peng Cheng Laboratory 
\fontsize{11pt}{\baselineskip}\selectfont
$^3$Yinwang Intelligent Technology Co. Ltd. \\
\begin{normalsize}${\tt zhanglk9@mail2.sysu.edu.cn} $\end{normalsize}
}

\begin{document}

\twocolumn[{%
\renewcommand\twocolumn[1][]{#1}%
\maketitle

}]

\renewcommand{\thefootnote}{}
\footnotetext{$^{\dagger}$ Co-corresponding author}

\begin{abstract}
Recent advances in Visual-Language-Action (VLA) models have shown promising potential for robotic manipulation tasks.
However, real-world robotic tasks often involve long-horizon, multi-step problem-solving and require generalization for continual skill acquisition, extending beyond single actions or skills. 
These challenges present significant barriers for existing VLA models, which use monolithic action decoders trained on aggregated data, resulting in poor scalability.
To address these challenges, we propose AtomicVLA, a unified planning-and-execution framework that jointly generates task-level plans, atomic skill abstractions, and fine-grained actions. 
AtomicVLA constructs a scalable atomic skill library through a Skill-Guided Mixture-of-Experts (SG-MoE), where each expert specializes in mastering generic yet precise atomic skills. 
Furthermore, we introduce a flexible routing encoder that automatically assigns dedicated atomic experts to new skills, enabling continual learning.
We validate our approach through extensive experiments. In simulation, AtomicVLA outperforms $\pi_{0}$ by 2.4\% on LIBERO, 10\% on LIBERO-LONG, and outperforms $\pi_{0}$ and $\pi_{0.5}$ by 0.22 and 0.25 in average task length on CALVIN. Additionally, our AtomicVLA consistently surpasses baselines by 18.3\% and 21\% in real-world long-horizon tasks and continual learning. These results highlight the effectiveness of atomic skill abstraction and dynamic expert composition for long-horizon and lifelong robotic tasks.
The project page is \href{https://zhanglk9.github.io/atomicvla-web/}{here}.
\end{abstract}

\section{Introduction}
\label{sec:intro}

Building on powerful Vision-Language Models~\cite{qwen2025qwen25technicalreport, deitke2024molmopixmoopenweights, beyer2024paligemmaversatile3bvlm, karamcheti2024prismaticvlmsinvestigatingdesign, wang2024qwen2vlenhancingvisionlanguagemodels, chen2023pali3visionlanguagemodels}, Vision-Language-Action (VLA) models~\cite{kim2024openvlaopensourcevisionlanguageactionmodel, black2024pi0visionlanguageactionflowmodel, brohan2023rt2visionlanguageactionmodelstransfer, liu2025rdt1bdiffusionfoundationmodel} unify visual perception, language understanding, and action generation into a single framework, achieving significant advances in robotic manipulation tasks. Despite this progress, current VLA models still face challenges in real-world deployments for complex long-horizon tasks and the continual acquisition of new skills. 

\begin{figure}[t]
    \centering
    \includegraphics[width=0.48 \textwidth]{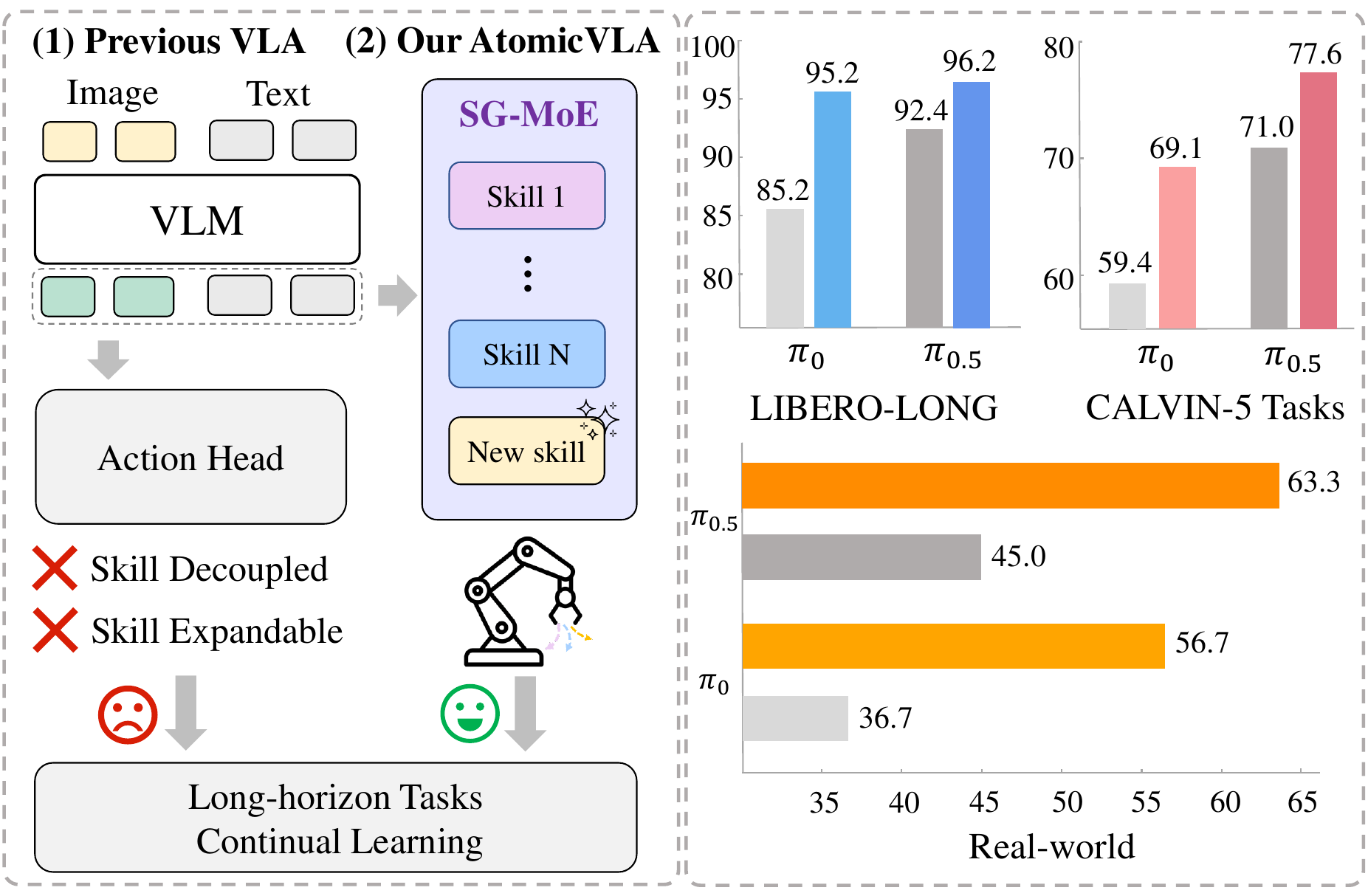} 
    \caption{\tb{Overview of AtomicVLA.} Unlike previous VLA models with a single action head, which suffer from limited scalability and severe interference among mixed skills, AtomicVLA employs a SG-MoE architecture to build a scalable skill expert library. By unifying task planning and action execution within this framework, it achieves strong performance on long-horizon and continual learning tasks in both simulation and real-world settings.} 
    \label{teaser} 
    \vspace{-14pt} 
\end{figure}

To overcome these challenges, a robotic model must support both high-level reasoning and fine-grained action generation, while enabling scalable continual learning.
To support high-level reasoning and task planning, some existing approaches employ a two-stage architecture~\cite{ahn2022icanisay, shi2025hirobotopenendedinstruction, geminiroboticsteam2025geminiroboticsbringingai, jiang2025galaxeaopenworlddatasetg0, nvidia2025gr00tn1openfoundation, hu2023lookleapunveilingpower}, where a pretrained vision-language model (VLM) serves as a high-level planner to generate subtask instructions, while a separate VLA-based controller translates these instructions into executable actions. 
However, recent studies~\cite{lin2025onetwovlaunifiedvisionlanguageactionmodel, yang2025lohovlaunifiedvisionlanguageactionmodel, yang2025instructvlavisionlanguageactioninstructiontuning} suggest that modular decoupling leads to a lack of mutual awareness between the planner and controller, causing suboptimal task coordination. Moreover, in real-world applications, this can result in the generation of outdated or irrelevant instructions due to system delays.
In addition, most existing VLA models rely on a single action-decoding module, limiting their scalability. Incrementally learning new skills requires fine-tuning existing models, which demands substantial computational resources and large datasets. Given the current scarcity of robot data, fully leveraging well-pretrained VLA model weights is essential during the scaling process. Moreover, when learning new skills incrementally, these models often interfere with previously acquired skills, leading to catastrophic forgetting and thereby hindering the lifelong learning capabilities.

To this end, we propose \name{},  as illustrated in Fig.~\ref{teaser}, an end-to-end framework that unifies task planning and action execution by adaptively generating either natural language instructions or latent actions. \name{} first infers the current execution state from the input observations and dynamically activates either its thinking module or its acting module. At task initialization or during transitions between sub-skills, the model triggers thinking to produce a task chain, create a task chain plan based on the current state, and outputs atomic skill abstractions. In the acting execution phase, it dynamically selects the corresponding skill-specific expert based on the most recent skill abstraction to generate precise robot control signals. 
Furthermore, to enable \name{} with continual learning capability, we introduce a Skill-Guided Mixture-of-Experts (SG-MoE) architecture that constructs a scalable library of atomic skills. This library comprises a shared expert and multiple dedicated skill experts, each focusing on mastering a specific atomic skill. Through a well-designed skill encoding mechanism and an extensible routing encoder, each atomic skill abstraction is mapped to a fixed embedding vector, allowing the routing module to rapidly adapt to new skills even as the skill library grows. When a new skill is introduced, only the corresponding expert and associated routing parameters need to be trained, leaving existing experts unchanged. This effectively prevents catastrophic forgetting, ensuring efficient and stable lifelong skill growth.

We conducted extensive experiments to validate the effectiveness of \name{} both in simulation platforms and real-world robots. In the LIBERO~\cite{liu2023liberobenchmarkingknowledgetransfer} benchmark, \name{} achieved an average performance improvement of 2.4\% over baseline models, with a notable 10\% improvement on the LIBERO-LONG. On the CALVIN~\cite{mees2022calvinbenchmarklanguageconditionedpolicy} benchmark, specifically on the task ABC-D training set, our method increased the average successful execution length by 0.22 and 0.25. Furthermore, we performed long-horizon task execution and continual learning experiments on a real-world Franka robot, where we observed performance improvements of 18.3\% and 21\%, respectively. These results further validate the potential of \name{}'s proposed atomic skill dynamic combination mechanism in supporting long-term task completion and lifelong skill accumulation. 
Overall, our contributions are as follows:
\begin{itemize}
    \item We introduce \name{}, an end-to-end framework that unifies task planning and action execution for long-horizon tasks and continual skill expansion. 
    \item We propose a Skill-Guided Mixture-of-Experts (SG-MoE) architecture and a scalable skill router for building a library of atomic skills.
    \item We validate the effectiveness of \name{} through extensive experiments conducted in both simulated environments and real-world robots.
\end{itemize}

\section{Related Work}
\label{sec:related work}

\subsection{Vision-Language-Action Models}

Vision-Language Action Models (VLAs) have emerged as a dominant paradigm in general-purpose robotic learning by leveraging the rich semantic priors and strong cross-modal generalization of large-scale Vision-Language Models (VLMs) pretrained on internet-scale data. Recent works \cite{kim2024openvlaopensourcevisionlanguageactionmodel,black2024pi0visionlanguageactionflowmodel,intelligence2025pi05visionlanguageactionmodelopenworld, zheng2025universalactionsenhancedembodied, brohan2023rt2visionlanguageactionmodelstransfer,pertsch2025fastefficientactiontokenization, wen2025dexvlavisionlanguagemodelplugin, lee2025molmoactactionreasoningmodels} fine-tune VLMs~\cite{deitke2024molmopixmoopenweights, beyer2024paligemmaversatile3bvlm, karamcheti2024prismaticvlmsinvestigatingdesign, wang2024qwen2vlenhancingvisionlanguagemodels} on diverse robotic datasets to directly map visual and linguistic inputs to motor actions, demonstrating impressive generalization to novel environments and tasks.  

However, constrained to some extent by the VLM's inherent hierarchical planning capability, most current VLAs exhibit limitations in structured task decomposition and long-horizon task planning. 
Several approaches introduce external high-level planners~\cite{erdogan2025plan,yang2025guiding,zhou2024isr, huang2024copageneralroboticmanipulation, geminiroboticsteam2025geminiroboticsbringingai, shi2025hirobotopenendedinstruction} that decompose long-horizon tasks into subgoals, which are then executed by a separate low-level policy. 
However, modular approaches often fail to unify action with vision and language in a shared latent space, resulting in misaligned decisions that compound loss.
To address this problem, recent work~\cite{yang2025lohovlaunifiedvisionlanguageactionmodel,lin2025onetwovlaunifiedvisionlanguageactionmodel,yang2025instructvlavisionlanguageactioninstructiontuning, chen2025fastinslowdualsystemfoundationmodel, liu2025hybridvlacollaborativediffusionautoregression, fang2025robixunifiedmodelrobot} propose integrated frameworks that jointly perform hierarchical reasoning and action generation within a unified model. 
Our work aligns with this direction: we adopt a Think-Act unified architecture, where a VLM simultaneously performs high-level task planning and atomic action abstraction, thereby directly guiding a specialized action expert to produce executable, temporally coherent action sequences.

\begin{figure*}[ht]
    \centering
    \includegraphics[width=1.0\textwidth]{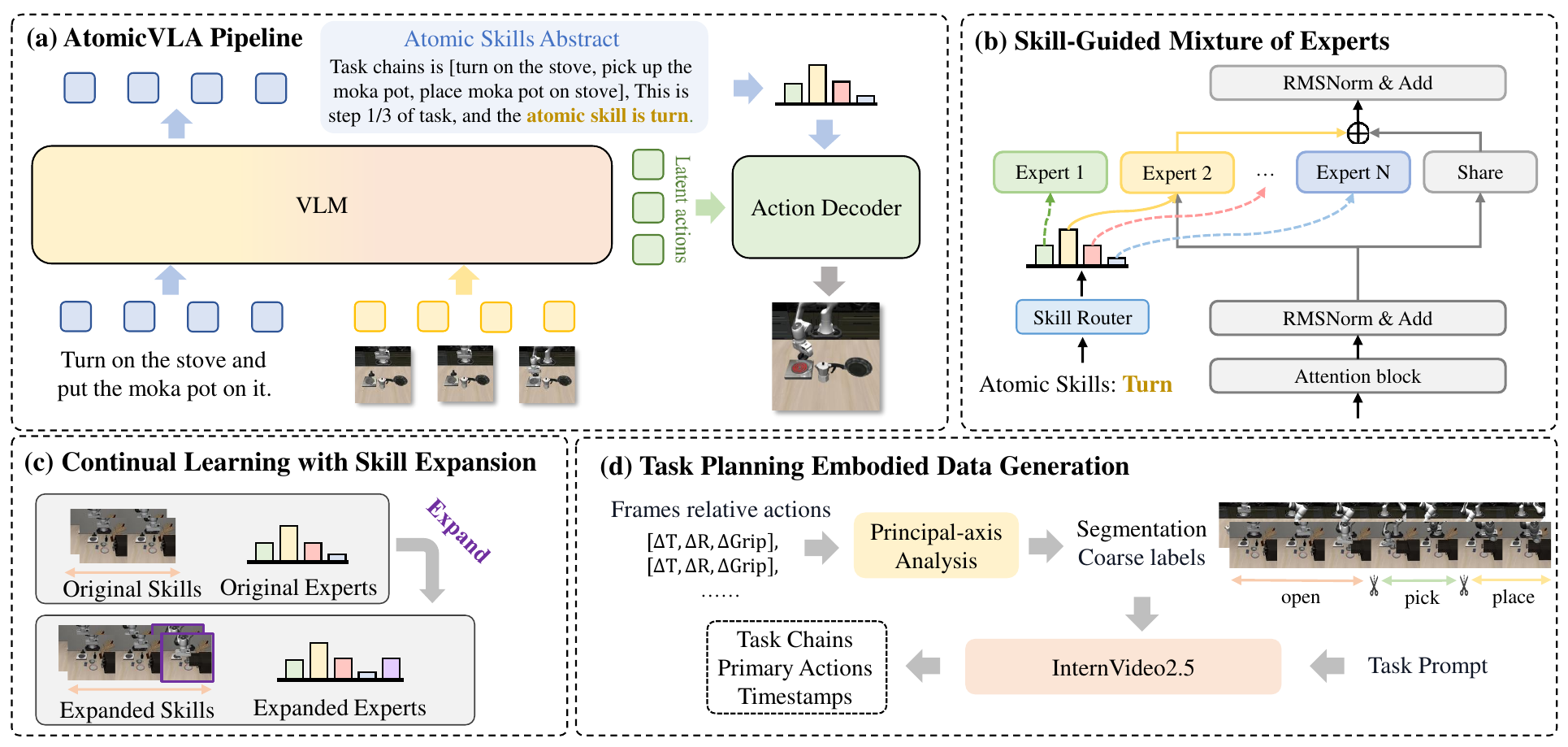} 
    \caption{\tb{(a) AtomicVLA Pipline.} AtomicVLA is a framework that unifies task planning and action execution. The VLM adaptively predicts atomic skill abstraction and latent action. Action Decoder in the SG-MoE architecture receives both the latent action and the newly inferred atomic skill abstraction, and generates fine grained motor actions. \tb{(b) Skill-Guided Mixture of Experts.} SG-MoE includes a skill router, a shared expert, and multiple atomic-skill experts. The router selects the top skill expert based on the atomic skill, and the action token is processed by both the activated skill expert and the shared expert. \tb{(c) Continual Learning with Skill Expansion.} New skills are added by training only the new expert and extending the router. \tb{(d) Task Planning Embodied Data Generation.} High-quality embodied reasoning data are generated using principal-axis analysis with InternVideo2.5~\cite{wang2025internvideo25empoweringvideomllms} model.} 
    \label{framework} 
    \vspace{-12pt} 
\end{figure*}

\subsection{Multimodal Mixture-of-Experts}

The Sparse Mixture-of-Experts (MoE) architecture has become a mainstream approach for scaling large language models (LLMs). 
By replacing the standard feed-forward layers with expert modules \cite{dai2024deepseekmoeultimateexpertspecialization,jiang2024mixtralexperts}, MoE improves task specialization and representation capability through conditional computation, while maintaining inference efficiency. 
In the field of autonomous driving, models such as \cite{yang2025drivemoemixtureofexpertsvisionlanguageactionmodel,xu2025moseskillbyskillmixtureofexpertslearning} design specialized MoE architectures tailored to multi-view observations and action skills, improving both trajectory prediction accuracy and inference efficiency.
Similarly, in robotics, some works \cite{huang2025mentormixtureofexpertsnetworktaskoriented,reuss2024efficientdiffusiontransformerpolicies,yang2025tramoelearningtrajectoryprediction, zhou2025chatvla2visionlanguageactionmodelopenworld, wang2025vervisionexperttransformer, yu2025forcevlaenhancingvlamodels} employ MoE to tackle task heterogeneity and long-tailed data distributions.
While these approaches demonstrate the utility of MoE for representation learning, they largely treat experts as interchangeable components within fixed architectural slots, without explicitly modeling structured, composable behaviors. In contrast, we reinterpret the MoE paradigm through the lens of skill modularity: we construct a dynamically scalable atomic skill library, where each expert corresponds to a semantically meaningful, reusable action primitive. Integrated with a pre-trained VLM that encodes atomic action abstractions, our approach enables a universal VLA model capable of both fine-grained skill decomposition and coherent long-horizon task composition.

\subsection{Continual Learning with Skill Abstractions}
To adapt to new tasks that emerge in dynamic environments, continual learning has become essential for developing general-purpose intelligent agents. Prior studies~\cite{wan2024lotuscontinualimitationlearning, mao2024dexskillsskillsegmentationusing,sun2025archhierarchicalhybridlearning, chitnis2020efficientbimanualmanipulationusing,fox2019hierarchicalvariationalimitationlearning,strudel2020learningcombineprimitiveskills} have leveraged unsupervised learning and hierarchical imitation learning to enable autonomous skill discovery from continuous data, which allows an agent to expand its skill set over time. Furthermore, to learn from streaming data without suffering from catastrophic forgetting, several approaches~\cite{lee2025incrementallearningretrievableskills,mete2024questselfsupervisedskillabstractions,yao2025thinksmallactbig,roy2025m2distillmultimodaldistillationlifelong} introduce latent action representations that abstract different skills and preserve previously acquired capabilities without relying on experience replay. Current VLA Models primarily focus on learning generalizable skills from broad pretraining, while dedicated investigations into continual learning remain limited. 
Although many VLAs~\cite{wen2025diffusionvlageneralizableinterpretablerobot,wang2025vqvlaimprovingvisionlanguageactionmodels} have explored various motion decoding methods, such as diffusion models~\cite{liu2025rdt1bdiffusionfoundationmodel}, flow matching~\cite{lipman2023flowmatchinggenerativemodeling}, and discrete encoding~\cite{oord2018neuraldiscreterepresentationlearning, mete2024questselfsupervisedskillabstractions}, they all use a single decoder. Their core focus is on the model's accuracy on the current task rather than its scalability.
We construct an expandable library of skill experts by using atomic units of robotic behavior together with a specialized routing module, which enhances the scalability of such models in skill acquisition.

\section{Method}
\label{sec:method}

\subsection{Overview}
As illustrated in Fig.~\ref{framework}, AtomicVLA integrates the thinking modality for task planning and the acting modality for action execution within a unified framework (Sec.~\ref{sec:3.2}). Building upon this architecture, we develop a skill-guided library of atomic action experts (Sec.~\ref{sec:3.3}) based on pi0 and introduce an extensible skill router that facilitates continual learning of new skills (Sec.~\ref{sec:3.4}) in real-world environments. To further ensure the generation of high-quality task planning data, we introduce an embodiment data generation pipeline (Sec.~\ref{sec:3.5}) grounded in principal axis analysis, which provides structured and consistent data to support effective task planning and execution.

\subsection{Unified Task Planning and Action Execution} 
\label{sec:3.2}

\mypara{Problem formulation.}
The central problem addressed in this section is to design a robot policy that simultaneously possesses task planning (thinking) and action execution (acting) capabilities, and can autonomously decide its output modality based on the current states. Specifically, in thinking mode, the policy takes multiple cameras observations $O_t^{1:n}$ and a language instruction $\ell$ as input and outputs a high-level task plan $[C_{0-k}, C_t,\sigma]$ in textual form. In contrast, in acting mode, the policy generates a concrete action command conditioned on the robot’s proprioceptive state $S_t$ and the most recent planning output $\sigma$.

\mypara{Adaptive thinking and acting.}
To enable seamless switching between the two output modalities, we introduce two special output tokens: $[think]$ and $[act]$. As illustrated in Algorithm ~\ref{algo:inference_pipeline}, given the current visual observations $O_t^{1:n}$ and task instruction $\ell$, the model first predicts identifier either $[think]$ or $[act]$. 
When the model outputs $[think]$, it enters the thinking mode, in which it generates a task chain $C_{0-k}$ that outlines the high-level plan, tracks the current execution progress $C_t$, and specifies the atomic skill abstraction $\sigma$ to be performed.
Typically, this mode is activated only at key time steps, such as task initiation or during the transition between sub-skills. Conversely, when $[act]$ is predicted, the model switches to acting mode, where it produces a low-level action chunk $A_t$ based on the atomic skill abstract $\sigma$ obtained in the most recent $[think]$ step and the current proprioceptive state.

\begin{figure}[t]
\centering
\begin{minipage}{1.0\columnwidth}
\begin{algorithm}[H]
\small
\begin{algorithmic}[1]
\Require VLA model $\pi_{\theta}$, language instruction $\ell$
\State $t \gets 0, \;O_{\text{t}}^{1:n} \gets \text{initial image}, \;Atomic \gets \text{none}$
\While {``task not done''}
    \State $M \sim \pi_{\theta}.\textsc{predict}(\cdot \mid O_{t}^{1:n}, \ell)$
    \If {$M = \texttt{[think]}$}
        \State $[C_{0-k}, C_t, \sigma] \sim \pi_{\theta}.\textsc{thinking}(\cdot \mid O_{t}^{1:n}, \ell)$
        \State $Atomic \gets \sigma \;$
    \ElsIf {$M = \texttt{[act]}$}
        \State $w_k \sim Router(embeded(Atomic))$
        \State $A_t \sim \pi_{\theta}.\textsc{acting}(\cdot \mid O_{t}^{1:n}, \ell, s_t, w_k)$
        \State Execute $A_t$
    \EndIf
    \State $t \gets t + 1$
\EndWhile
\end{algorithmic}
\caption{Inference Pipeline of AtomicVLA}
\label{algo:inference_pipeline}
\end{algorithm}
\end{minipage}
\vspace{-10pt} 
\end{figure}

\subsection{Skill-guided Mixture of Experts Architecture}
\label{sec:3.3}
\mypara{Atomic skill abstract embedding.} 
To enhance the representational distinctiveness among atomic skills, we adopt an encoding strategy inspired by noise scheduling in diffusion-based denoising models. Specifically, each atomic skill abstract is mapped to a scalar noise level \(\sigma \in [0, 100]\), which is then embedded into a high-dimensional vector. This continuous and structured embedding space facilitates semantic separation across skills and enables robust routing to the corresponding skill-specific experts.
\begin{equation}
Z_{\sigma} = E(norm(\log(\sigma))),
\end{equation}
where \(\sigma\) denotes the assigned noise level for the skill, and \(E(\cdot)\) is a embedding function that maps the normalized scalar to a high-dimensional embedding vector \(Z_{\sigma}\).

\mypara{Skill-Guided dynamic routing.}
We build upon the $\pi_{0}$ vision-language-action (VLA) foundation model, a generalist robotic policy pretrained on large-scale multimodal data, and extend it with an atomic action abstraction-guided Mixture-of-Experts (MoE) architecture to construct a scalable atomic skill library. As illustrated in Fig.~\ref{framework}(b), our skill library consists of three key components: (1) a skill router, (2) a shared expert that maintains the pre-trained action generation capabilities of $\pi_{0}$, and (3) multiple atomic skill experts, each specialized in executing a distinct atomic skill.

To maintain the specialized skills of individual atomic experts, we first derive an atomic action abstraction from the high-level task instruction and environmental observation via thinking pipeline. This abstraction is deterministically mapped to a fixed high-dimensional embedding \(Z_\sigma \in \mathbb{R}^d\), which serves as the conditioning signal for the skill router. The router computes a probability distribution over experts as:
\begin{equation}
w_{k} = \text{Router}(Z_\sigma), \quad k \in \{1, 2, \dots, K\},
\end{equation}
where \(K\) denotes the number of atomic skill experts. We adopt a sparse activation strategy: only the top-scoring expert is selected for action generation. Let $k$ be the index of the activated expert, and let its raw score be $w_{k}$. The final action chunk \(A_t\) is computed as a weighted combination of the shared expert and the selected atomic expert:
\begin{equation}
{F}_{out} = (1 - w_{k}) \cdot {F_{share}}({x}_t) + w_{k} \cdot {F}_{k}({x}_t),
\end{equation}
where \(\mathbf{x}_t\) denotes the current multimodal input $[O_{t}^{1:n}, \ell, s_t]$. This architecture enables the system to retain the strong generalization capability of $\pi_{0}$ while achieving high-fidelity execution of specific skills through dedicated experts.

\begin{figure*}[t]
    \centering
    \includegraphics[width=1.0\textwidth]{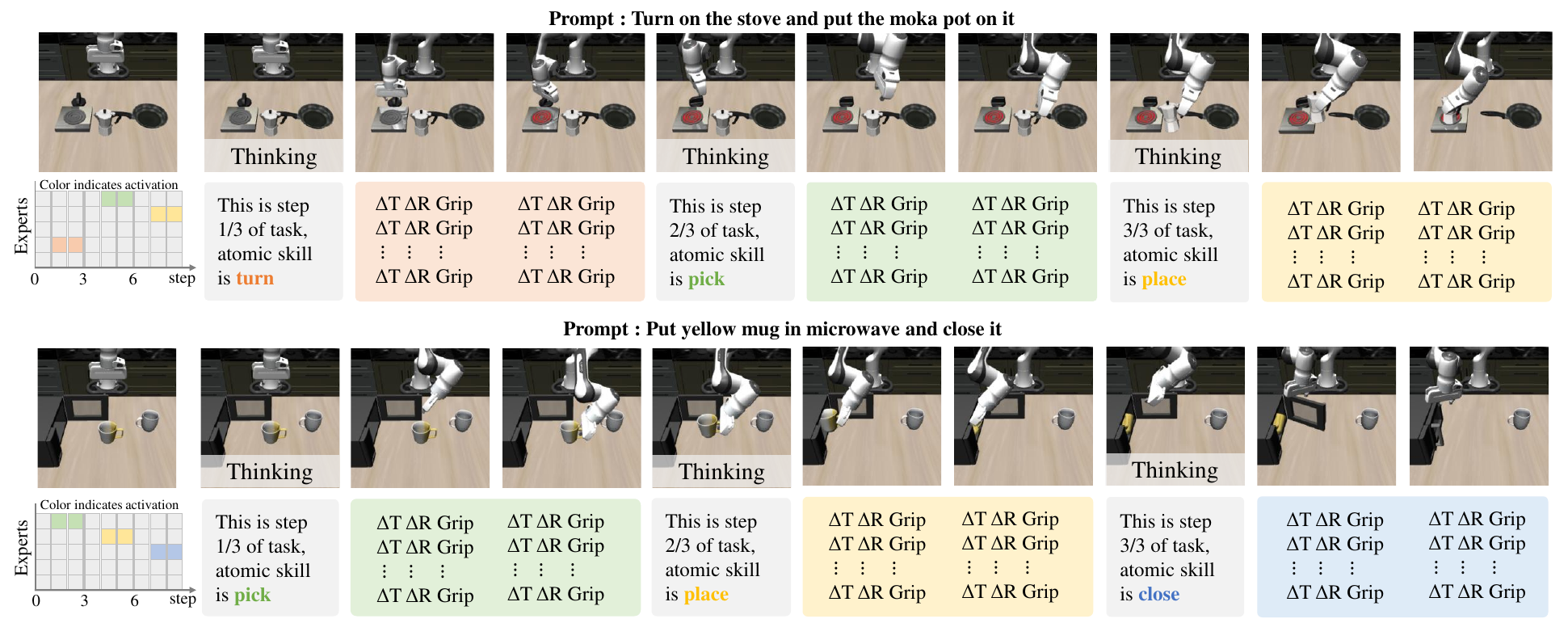} 
    \caption{\tb{Inference Example of AtomicVLA.} We visualize two tasks from LIBERO-LONG. For each task, the top row shows the task progression, and the bottom row shows AtomicVLA’s inferred outputs. Gray blocks denote Thinking, while colored blocks indicate Acting, with colors corresponding to the activated skill experts.
    The left row shows the initial task state (top) and the skill-expert activation during inference (bottom).} 
    \label{example——libero} 
    \vspace{-12pt} 
\end{figure*}

\subsection{Continual Learning with Skill Expansion}
\label{sec:3.4}
In real-world deployments, robots inevitably encounter new tasks that require atomic skills not previously observed during training. Directly incorporating these novel skills into the existing skill library and retraining the entire model often leads to catastrophic forgetting, significantly impairing the performance of previously learned skills.

AtomicVLA adopts a modular skill-expert mechanism, which enables continual scalability of the skill library. Specifically, as introduced in Sec.~\ref{sec:3.3}, each atomic skill is mapped to a fixed high-dimensional embedding vector \({Z}_\sigma\), providing an explicit semantic abstraction of the skill. This design inherently enables incremental learning in lifelong settings: when a new atomic skill is introduced, it is sufficient to add a corresponding expert module to the existing architecture and extend the routing network. 

To ensure smooth integration, the expanded router is initialized by copying weights from the original router, while the new routing branch is initialized with small random values. This initialization strategy allows the model to adapt to the enlarged skill set with minimal fine-tuning, while preserving the performance of previously acquired skills. Consequently, AtomicVLA achieves efficient and interference-free expansion of its atomic skill library, a crucial requirement for scalable lifelong robotic learning.

\subsection{Task Planning Embodied Data Generation}
\label{sec:3.5}

To obtain accurate and reliable annotations of atomic actions, we propose a trajectory-based atomic decomposition method grounded in principal-axis analysis. Traditional approaches often rely on Vision-Language Models for video understanding or optical flow-based motion features to segment action sequences. However, these methods are prone to ambiguity and noise, which typically require extensive manual post-processing to correct and refine the results.

In contrast, our method analyzes the key kinematic dimensions of the end-effector trajectory, including translational displacements (\(\Delta x, \Delta y, \Delta z\)), rotational changes (\(\Delta \text{roll}, \Delta \text{pitch}, \Delta \text{yaw}\)), and binary gripper states, to achieve coarse but semantically meaningful segmentation of atomic actions. Specifically, for each short motion chunk, we identify the dominant mode of motion by comparing the magnitudes of translational and rotational components. Concurrently, gripper state transitions are tracked to infer action semantics and execution progress. For instance, a continuous decrease in the \(z\)-coordinate combined with a gripper closing event indicates a “pick” action, whereas limited translational movement accompanied by significant rotation with a closed gripper is classified as a “turn” operation. This physics-informed decomposition produces temporally precise and semantically interpretable boundaries for atomic actions, substantially reducing the reliance on manual refinement.

Based on the output of principal-axis analysis, we decompose a full task trajectory into a temporally ordered sequence of atomic action segments. To refine and validate the semantic labels of these segments, we employ the InternVideo2.5 model~\cite{wang2025internvideo25empoweringvideomllms} to interpret the corresponding video clips, enabling automatic correction and enrichment of the initial atomic action annotations. By aligning these refined labels with the full trajectory, we construct a structured reasoning chain comprising the sequence of executed atomic actions and the associated high-level plan for subsequent steps. This integrated representation not only improves the fidelity of atomic action annotation but also provides interpretable, step-by-step execution guidance that supports robust long-horizon task planning and decision-making.

\begin{table*}[t]
\centering
\small
\caption{\textbf{Comparison of Different Methods on LIBERO Benchmark(\%).}}
\label{tab:libero}
\vspace{-3pt}
\addtolength{\tabcolsep}{13.2pt}
\begin{tabularx}{0.84\linewidth}{lccccc}
\toprule
\tb{Method} & \tb{Spatial} & \tb{Object} & \tb{Goal} &  \cellcolor{myblue}{\tb{Long}} & \tb{Avg.} \\
\midrule
Octo \cite{octomodelteam2024octoopensourcegeneralistrobot}      & 78.9  & 85.7  & 84.6  & \cellcolor{myblue}{51.1}  & 75.1  \\
OpenVLA \cite{kim2024openvlaopensourcevisionlanguageactionmodel}& 84.9  & 88.4  & 79.2  & \cellcolor{myblue}{53.7} & 76.5  \\
SpatialVLA \cite{qu2025spatialvlaexploringspatialrepresentations} & 88.2  & 89.9  & 78.6  & \cellcolor{myblue}{55.5} & 78.1  \\
CoT-VLA \cite{zhao2025cotvlavisualchainofthoughtreasoning} & 87.5  & 91.6  & 87.6  & \cellcolor{myblue}{69.0} & 83.9  \\
$\pi_{0}$~\cite{black2024pi0visionlanguageactionflowmodel} & 96.4  & \tb{98.8} & 95.8  & \cellcolor{myblue}{85.2}  & 94.2  \\
$\pi_{0.5}$~\cite{intelligence2025pi05visionlanguageactionmodelopenworld} & \tb{98.8}  & 98.2  & \tb{98.0}  & \cellcolor{myblue}{92.4}  & 96.9\\
\midrule
\rowcolor{mygray}  \tb{\name{} (Ours)}       & 96.8  & 98.0  & 96.4  & \cellcolor{myblue}{\tb{95.2}} & 96.6 \\

\rowcolor{mygray}  \tb{\name{}* (Ours)}    & \tb{98.8} & \tb{98.8} & 97.2  & \cellcolor{myblue}{\tb{96.2}} & \tb{97.8} \\
\bottomrule
\end{tabularx}
\vspace{-2pt}
\end{table*}
\begin{table*}[t]
\centering
\small
\caption{\textbf{Long-horizon Robotic Manipulation Evaluation on CALVIN Benchmark(\%).}}
\label{tab:calvin}
\vspace{-3pt}

\addtolength{\tabcolsep}{5.5pt}
\begin{tabularx}{0.84\linewidth}{l l ccccc c}
\toprule

 \multirow{2}{*}{\tb{Method}} &  \multirow{2}{*}{\tb{Task}} & \multicolumn{5}{c}{\tb{Tasks Completed in a Row}} &  \multirow{2}{*}{\tb{Avg. Len $\uparrow$}} \\
\cmidrule(lr){3-7}
 &  & 1 & 2 & \cellcolor{myblue}{3} & \cellcolor{myblue}{4} & \cellcolor{myblue}{5} &  \\
\midrule

$\pi_{0}$~\cite{black2024pi0visionlanguageactionflowmodel}  & ABC$\to$D & 94.3 & 87.0 & \cellcolor{myblue}{77.9} & \cellcolor{myblue}{68.5} & \cellcolor{myblue}{59.4} & 3.87 \\
$\pi_{0.5}$~\cite{intelligence2025pi05visionlanguageactionmodelopenworld} & ABC$\to$D & 91.9 & 84.6 & \cellcolor{myblue}{79.4} & \cellcolor{myblue}{75.5} & \cellcolor{myblue}{71.0}& 4.02 \\

\midrule
\rowcolor{mygray}  \textbf{\name{} (Ours)}    & ABC$\to$D  & 95.0 & 87.8 & \cellcolor{myblue}{\textbf{81.9}} & \cellcolor{myblue}{\textbf{75.0}} & \cellcolor{myblue}{\textbf{69.1}} & \textbf{4.09} \\

\rowcolor{mygray}  \textbf{\name{}* (Ours)}    & ABC$\to$D  & 94.1 & 88.7 & \cellcolor{myblue}{\textbf{85.2}} & \cellcolor{myblue}{\textbf{81.7}} & \cellcolor{myblue}{\textbf{77.6}} & \textbf{4.27} \\

\bottomrule
\end{tabularx}
\vspace{-6pt}
\end{table*}


\vspace{-5pt}
\section{Experiments}
\label{sec:experiments}

\subsection{Experiments Setup}
\label{sec: setup}

\mypara{Benchmarks.}
We evaluate AtomicVLA and AtomicVLA* on two widely adopted robotic manipulation benchmarks: LIBERO \cite{liu2023liberobenchmarkingknowledgetransfer} and CALVIN \cite{mees2022calvinbenchmarklanguageconditionedpolicy}. For the LIBERO benchmark, we assess model performance across all four task suites.
To further examine the model’s capability in long-horizon planning and compositional generalization, we perform additional experiments on the CALVIN benchmark using the ABC-D split.

\mypara{Training setup.}
We build AtomicVLA and AtomicVLA* upon the pretrained \(\pi_0\) and \(\pi_{0.5}\) foundation model. The models were trained using robot trajectory data formatted according to the Lerobot standard. 
We use 5 skill experts for both the LIBERO benchmark suite and real-world robot experiments. For the CALVIN benchmark, we employ 8 skill experts to cover its broader task vocabulary. 
Further implementation details are provided in the Appendix.

\mypara{Real-world robot.}
We conduct real-world experiments using a Franka robotic arm, which includes three long-horizon tasks and five different types of short tasks. For each short-horizon task, we collect 50 trajectories, while each long-horizon task contains 100 trajectories, resulting in a total of 550 real-world demonstration trajectories. The five short tasks cover different categories of manipulation actions, including Grasp block, Stack blocks, Close microwave, Press button, and Open drawer. The long-horizon tasks include:

\begin{itemize}
    \item \tb{Objects in plate}: place all blocks on the table into a green plate.
    \item \tb{Object into drawer}: open the top drawer and place the block inside.
    \item \tb{Object into microwave}: place the plate into the microwave and close the door.
\end{itemize}

\begin{figure}[t]
    \centering
    \includegraphics[width=0.47\textwidth]{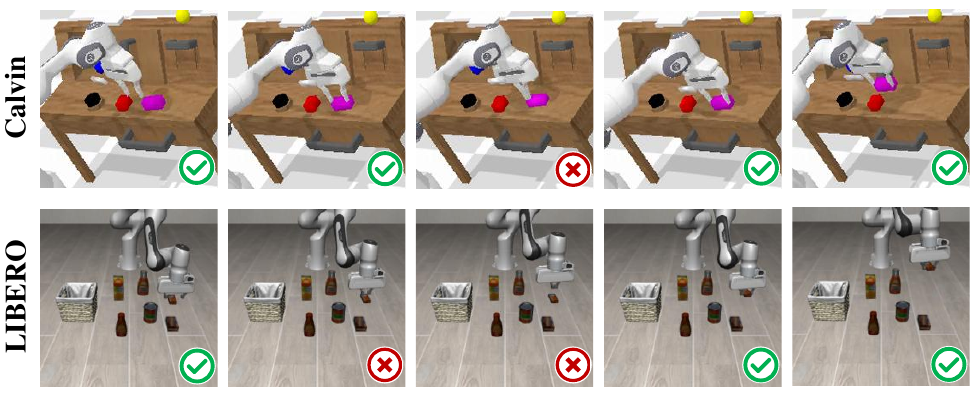} 
    \caption{\tb{Error Recovery Capability Demonstration.} When encountering a skill execution failure, AtomicVLA automatically assesses the progress and re-executes the current skill.} 
    \label{fixes} 
    \vspace{-12pt} 
\end{figure}

\begin{figure*}[htbp]
    \centering
    \includegraphics[width=1.0\textwidth]{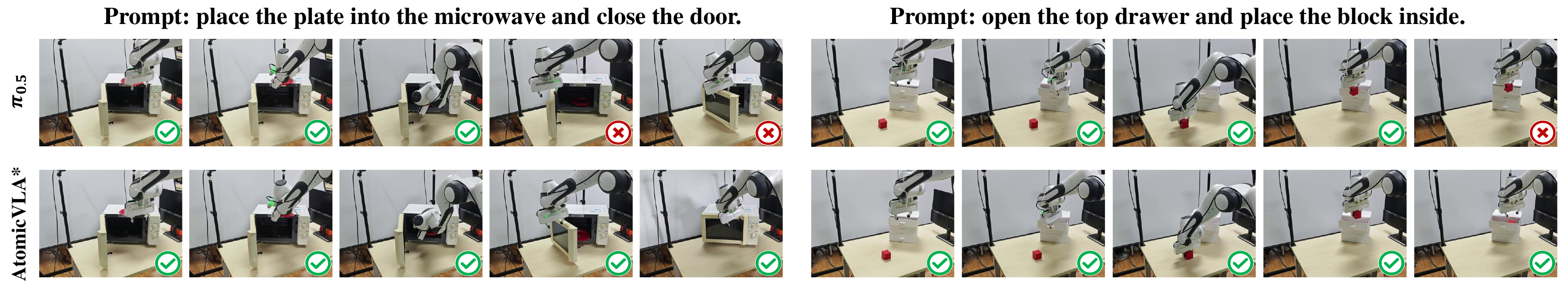} 
    \caption{Demonstrations show the execution process of AtomicVLA* (second row) and baselines $\pi_{0.5}$ (first row).} 
    \label{real-world} 
\vspace{-6pt}
\end{figure*}

\subsection{Results on Simulation}
\label{sec:sim}

\mypara{Results on LIBERO.}
As shown in Tab.~\ref{tab:libero}, AtomicVLA achieves an average success rate of 96.6\%  across the four suites, outperforming the strong baseline by 2.4\%. Notably, on the most challenging LIBERO-LONG suite, AtomicVLA attains a success rate of 95.2\%, representing a 10\% improvement over the $\pi_0$. Furthermore, AtomicVLA* demonstrates even stronger performance, reaching an average success rate of 97.8\% and 96.2\% on LIBERO-LONG.

This superior performance can be attributed to the core mechanism of AtomicVLA, which explicitly decomposes long-horizon tasks into a sequence of atomic skill abstractions and dynamically activates the corresponding skill experts. The ``decompose–plan–compose” paradigm naturally aligns with the structure of multi-stage robotic tasks. As illustrated in Fig.~\ref{example——libero}, at the beginning of each atomic subtask, AtomicVLA generates a precise skill-level action abstraction to guide the selection of the appropriate expert. Importantly, when an execution failure occurs, for example, the butter is grasped but subsequently dropped as illustrated in Fig.~\ref{fixes}, AtomicVLA can detect the task anomaly, regenerate a new atomic skill abstraction, and recover from the error to resume task execution.

\mypara{Results on Calvin.} 
As shown in Tab.~\ref{tab:calvin}, AtomicVLA achieves an average task length of 4.09, outperforming the $\pi_{0}$ baseline by 0.22, while AtomicVLA* reaches an average task length of 4.27, outperforming the $\pi_{0.5}$ baseline by 0.25. Notably, AtomicVLA* demonstrates superior overall task completion rate with relative improvements of 5.8\%, 6.2\%, and 6.6\%  on the last three stages of the evaluation sequence. These results indicate that AtomicVLA is particularly effective in handling temporally extended and sequential manipulation tasks.

As illustrated in Fig.~\ref{fixes}, we also observe that AtomicVLA exhibits a capability for error recovery in experiments. However, due to the evaluation constraints of the CALVIN benchmark, successful recoveries after failures are not considered valid completions, which prevents subsequent tasks from being executed. As a result, the reported performance metrics may slightly underestimate the true capability of the model.

\begin{table}[t]
\centering
\small
\caption{\tb{Long-horizon Multi-task Experiments(\%).} InP, IntoD, and IntoM stand for Objects in plate, Object into drawer, Object into microwave, respectively.}
\label{tab:real}
\vspace{-3pt}
\addtolength{\tabcolsep}{0.1pt} 
\begin{tabularx}{0.98\linewidth}{lccccc}
\toprule
\tb{Method}   & \tb{InP} &  \tb{IntoD} & \tb{IntoM} & \tb{Avg.} & \tb{$\Delta$Avg.} \\
\midrule 
$\pi_{0}$~\cite{black2024pi0visionlanguageactionflowmodel}          & 45 & 55 & 10 & 36.7 &  – \\
$\pi_{0.5}$~\cite{intelligence2025pi05visionlanguageactionmodelopenworld}        & 65 & 35 & 35 & 45   &  –  \\
\midrule 
\rowcolor{mygray} \textbf{\name{}}     & 65 & \tb{60} & 45 & 56.7 & \cellcolor{myblue}{$+20.0\uparrow$}  \\
\rowcolor{mygray} \textbf{\name{}*}    & \tb{75} & \tb{60} & \tb{55} & \tb{63.3} & \cellcolor{myblue}{${+18.3}\uparrow$} \\
\bottomrule
\end{tabularx}
\vspace{-3pt}
\end{table}

\subsection{Results on Real-world Robot}
\label{sec:real}
\mypara{Long-horizon Tasks.}
We perform mixed training using the collected data from three long-horizon tasks.
As shown in Tab.~\ref{tab:real}, AtomicVLA and AtomicVLA* outperform the baseline model by 20\% and 18.3\%, respectively. As illustrated in Fig.~\ref{real-world}, we present two representative long-horizon tasks. AtomicVLA* reliably completes the experimental configurations that $\pi_{0.5}$ fails to accomplish, and this advantage becomes more evident in tasks involving door-closing operations. Building on this observation, AtomicVLA* demonstrates stronger robustness and execution stability across complex manipulation sequences. 

\begin{figure}[t]
    \centering
    \includegraphics[width=0.47\textwidth]{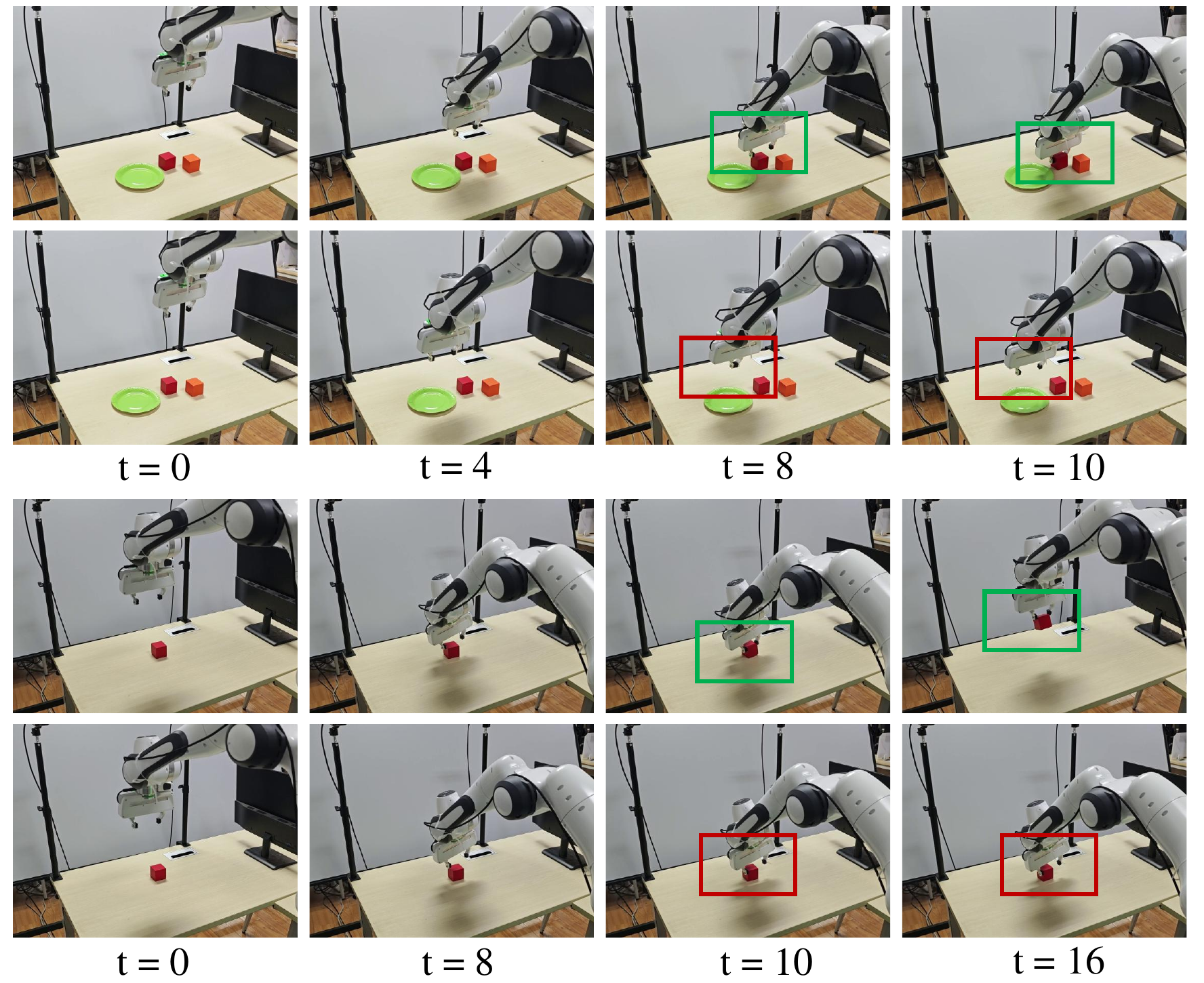} 
    \caption{\tb{Mixed-Training Skill Interference and Continual-Learning Degradation.} The top two rows illustrate skill interference in long-horizon tasks: the first shows successful single-skill executions, while the second shows failures after mixed training. The bottom two rows show degradation after continual learning: the first row presents the performance of $\pi_{0.5}$ before learning new skills, and the second shows its performance afterward. Red and green boxes highlight the key differences.
    } 
    \label{mix} 
    \vspace{-12pt} 
\end{figure}

Previous real-world studies on robotic manipulation typically focus on training and evaluating a single specific task, while joint training across multiple heterogeneous tasks has been relatively uncommon. Our observations indicate that combining tasks with large differences can lead to mutual interference, which in turn limits overall performance. This effect becomes particularly pronounced in tasks that involve significant changes in gripper states across different execution stages. For instance, in the ``Object into drawer" task, the drawer-opening subtask does not require gripper closure, which can adversely affect the model’s behavior on other grasping-related tasks, resulting in unintended gripper opening or closing actions, as illustrated in Fig.~\ref{mix}. By constructing an explicit library of atomic skills, AtomicVLA effectively mitigates such cross-task interference. Each skill precisely activates its corresponding expert to execute the required operation, which substantially alleviates the interference between heterogeneous skills and overcomes the performance bottleneck of mixed multi-task training.

\begin{table*}[ht!]
\centering
\small
\caption{\tb{Continual Learning with Skill Expansion(\%).} $\Delta$Avg. represents the average performance change on the four base tasks after learning new skills compared with their performance before learning.CL is continual learning}
\label{tab:lifelong}
\vspace{-3pt}
\addtolength{\tabcolsep}{7.2pt}
\begin{tabularx}{0.92\linewidth}{lccccccc} 
\toprule
\tb{Method}        & \tb{Grasp}  &\tb{Stack} & \tb{Close} &\tb{Press} & \cellcolor{myblue}{\tb{Open (new)}}  & \tb{Avg.} & \tb{$\Delta$Avg.}\\
\midrule
$\pi_{0.5}$~\cite{intelligence2025pi05visionlanguageactionmodelopenworld}          &     85    &   65    &   70    &   90    &   \cellcolor{myblue}{-}   &  77.5 & – \\
$\pi_{0.5}$~\cite{intelligence2025pi05visionlanguageactionmodelopenworld} (CL)  &   70     &    45    &   60    &   75   &   \cellcolor{myblue}{55}   & 61 & \cellcolor{mypink}{-$15.0\downarrow$} \\
\rowcolor{mygray} \textbf{\name{}*}  &    \tb{95}    &    \tb{80}    &   \tb{70}    &   \tb{100}   &   \cellcolor{myblue}{-}   & \tb{86.3} & – \\
\rowcolor{mygray}  \textbf{\name{}*} (CL) &    90   &   \tb{80}    &   \tb{80}    &  \tb{100}   &   \cellcolor{myblue}{\tb{70}}         & \tb{82} & \cellcolor{mypink}{\tb{$-1.3\downarrow$}} \\
\bottomrule
\end{tabularx}
\vspace{-5pt}
\end{table*}

\mypara{Continual learning skills.}
To evaluate the effectiveness of our proposed lifelong skill expansion mechanism in real-world scenarios, we conduct training and evaluation on short-horizon task dataset consisting of five diverse manipulation categories. In this experiment, the ``open" operation is treated as a new atomic skill, which is introduced as an additional capability during the continual learning phase after the initial training stage. Specifically, we first perform mixed training on four short-horizon tasks and train the ``open" skill independently on top of the pretrained model.

Learning a new skill often causes substantial interference with previously acquired abilities in conventional baseline models, leading to noticeable performance degradation. As illustrated in Fig.~\ref{mix}, in a case that was originally expected to succeed, the task could not be completed after continual learning. The gripper failed to close promptly after reaching the target position. As shown in Tab.~\ref{tab:lifelong}, the average success rate of $\pi_{0.5}$ decreases by approximately 15\%, with the stack task exhibiting the most severe interference, showing a 20\% decrease. In contrast, AtomicVLA* maintains stable performance after continual learning. Owing to its structured skill library management, the previously learned skills remain largely unaffected. Moreover, under the same number of training steps, AtomicVLA* acquires new skills more efficiently and achieves an overall improvement of 21\% across all five tasks compared to $\pi_{0.5}$. These findings highlight our effectiveness for continual learning.

\subsection{Ablation Study}
\label{sec:ablation}
We conduct ablation experiments on the LIBERO-LONG benchmark to evaluate the effectiveness of our skill-aware routing mechanism. Specifically, we compare AtomicVLA against three baselines: (i) a non-MoE \(\pi_0\)-based baseline, (ii) a standard token-level Mixture-of-Experts (MoE) that selects experts independently for each action token, and (iii) a variant adapted from MoDE~\cite{reuss2024efficientdiffusiontransformerpolicies}, which conditions expert selection on the denoising timestep \(t\) (i.e., using \(t\) as the routing signal). 

\vspace{-3pt}
\begin{table}[ht!]
\centering
\small
\caption{\tb{Results on LIBERO Benchmark(\%).}}
\label{tab:libero10}
\vspace{-3pt}
\addtolength{\tabcolsep}{13.9pt}
\begin{tabularx}{0.9\linewidth}{lc}
\toprule
\tb{Method}        & \tb{LIBERO-LONG} \\
\midrule
$\pi_{0}$~\cite{black2024pi0visionlanguageactionflowmodel}          &     85.2      \\
+ MoE           &      88.6     \\
+ MoDE~\cite{reuss2024efficientdiffusiontransformerpolicies}  &       89.5    \\
\rowcolor{mygray}  \textbf{+ SG-MoE (Ours)}  &      \tb{95.2}   \\
\bottomrule
\end{tabularx}
\vspace{-8pt}
\end{table}

As shown in Tab.~\ref{tab:libero10}, AtomicVLA achieves a success rate of 95.2\%, outperforming the MoE baseline by 6.6\% and the timestep-conditioned MoDE variant by 5.7\%.  The experimental results indicate that the performance gap between the MoE-based and MoDE-based methods is relatively small. This is primarily because both approaches rely on token-level expert routing, where the improvements largely stem from load balancing that distributes tokens across experts. As a result, each expert still learns a mixture of skills without clear specialization. In contrast, SG-MoE employs atomic skill abstractions as the routing criterion, which ensures that all tokens associated with a specific skill stage are consistently processed by the corresponding expert network. Consequently, each expert focuses on a single skill with a similar action distribution, reducing interference among different skills. Moreover, this notable performance gain demonstrates that routing experts based on semantically meaningful atomic skills, rather than on individual action tokens or denoising steps, leads to more coherent and efficient skill execution in long-horizon tasks.

\section{Conclusion}
\label{sec:conclusion}
In this paper, we introduce AtomicVLA, an end-to-end framework that unifies task planning and action execution for long-horizon tasks and continual skill expansion.  We design a unified architecture capable of adaptively deciding task plans and generating latent action outputs, and construct an atomic skill-guided expert library based on our proposed SG-MoE architecture and the specialized skill router. AtomicVLA is inherently scalable: when learning new skills, it only requires extending the skill router and adding the corresponding new skill experts to rapidly acquire the novel capabilities. We validate AtomicVLA in both simulated and real-world robotic environments, demonstrating its superior performance in long-horizon tasks and continual learning. Notably, it effectively mitigates skill interference arising from joint training and alleviates knowledge forgetting and performance degradation during continual skill acquisition, highlighting its significant potential for scalable continual learning in vision-language-action models.

\section*{Acknowledgments}
This work was supported by the National Key Research and Development Program of China (2024YFE0203100), the Scientific Research Innovation Capability Support Project for Young Faculty (No. ZYGXQNJSKYCXNLZCXM-I28), the National Natural Science Foundation of China (NSFC) under Grants No. 62476293, No. 62372482 and No. 62272494, and in part by the Major Key Project of PCL (Grant No. PCL2025A17) and the General Embodied AI Center of Sun Yat-sen University.

{
    \small
    \bibliographystyle{ieeenat_fullname}
    \bibliography{main}
}

\clearpage

\maketitlesupplementary

\begingroup

\setcounter{tocdepth}{2}

\startcontents[supplements]

\renewcommand{\thesection}{A.\arabic{section}}
\renewcommand{\thesubsection}{\thesection.\arabic{subsection}}

\setcounter{section}{0}

\setlength{\cftbeforesecskip}{0.8em} 
\cftsetindents{section}{0em}{2.5em}
\cftsetindents{subsection}{1em}{3.3em}

\etoctoccontentsline{part}{Appendix}
\localtableofcontents


\section{Video Demonstration}
Please refer to the video file in the attachment for a quick overview of AtomicVLA.

\section{Future Work and Limitations}
Most current vision-language-action (VLA) models are typically trained and evaluated on individual tasks. In this work, we investigate skill interference arising from multi-skill joint training through controlled experiments and introduce a Skill-Gated Mixture-of-Experts (SG-MoE) framework to construct a scalable atomic skill library, thereby exploring the potential of VLA models in long-horizon tasks and continual learning. Although this paradigm shows clear promise, many advantages remain insufficiently explored.

\begin{itemize}
    \item AtomicVLA relies on a task planning module that produces accurate atomic skill abstractions and on a set of well trained skill experts. The skill router relies on the VLM to produce accurate atomic skill abstractions during task execution, a capability constrained by the VLM’s reasoning and planning fidelity. Recent studies such as Embodied Brain~\cite{ji2025robobrainunifiedbrainmodel,tan2025reasonrftreinforcementfinetuningvisual, baairobobrainteam2025robobrain20technicalreport} and $\pi_{0.5}$~\cite{intelligence2025pi05visionlanguageactionmodelopenworld} indicate that combining large scale web data with embodied experience can effectively train VLMs that are capable of skill decomposition and task planning while enabling the construction of a high quality expert skill library, which can further enhance the performance of AtomicVLA. 
    \item By decoupling skill learning, AtomicVLA substantially mitigates interference during multi-skill training and demonstrates strong adaptability to new skills. However, acquiring new tasks still requires collecting substantial human demonstration data for imitation learning(IL). Notably, recent works like $\pi^{*}_{0.6}$~\cite{intelligence2025pi06vlalearnsexperience}, SimpleVLA-RL~\cite{li2025simplevlarlscalingvlatraining} and VLA-RL~\cite{lu2025vlarlmasterfulgeneralrobotic} have shown that reinforcement learning(RL) can effectively train VLA models and achieve strong performance. Integrating a pre-trained skill expert library with reinforcement learning(RL) may empower AtomicVLA to generalize to novel tasks under few-shot or even zero-shot settings.
\end{itemize}



\section{Additional Details}
\label{sec:details}

\subsection{Training Setup}
\label{subsec:setup_supp}
For all experiments, we construct the skill library using one shared expert together with multiple skill experts. Each skill expert follows the Gemma architecture, where the feedforward module is implemented with an independent SwiGLU activated MLP. All skill experts are randomly initialized at the beginning of training to enable disentangled skill representations and support incremental learning. The model configuration is width = 2048, mlp dim = 4096, depth = 18, num heads = 8, and head dim = 256. Building on this configuration, the learning rate follows  $CosineDecaySchedul$ with a warm up phase of 1,000 steps, a peak learning rate of \(2.5 \times 10^{-5}\), and a final learning rate of \(5 \times 10^{-6}\). The optimizer is AdamW with a gradient clipping norm = 1.0.To stabilize training, an exponential moving average (decay = 0.999) is used throughout optimization.

Following this setup, we train the model for 100k iterations on both the LIBERO and Calvin simulation platforms, and for 30k iterations in real world robotic experiments, with a batch size of 64. All training is performed on 8 $\times$ H200 GPUs, and inference is conducted on a single NVIDIA RTX RPO6000 GPU. 

\begin{table}[t]
\centering
\caption{\tb{Atomic skill distribution in the LIBERO dataset.}}
\begin{tabular}{l c}
\toprule
\textbf{Atomic Skill} & \textbf{Count} \\
\midrule
Pick  & 2462 \\
Place & 761  \\
Open  & 201  \\
Close & 152  \\
Turn  & 175  \\
\bottomrule
\end{tabular}
\label{tab:libero_atomic_skill_distribution}
\end{table}

\begin{figure*}[t]
    \centering
    \includegraphics[width=1.0\textwidth]{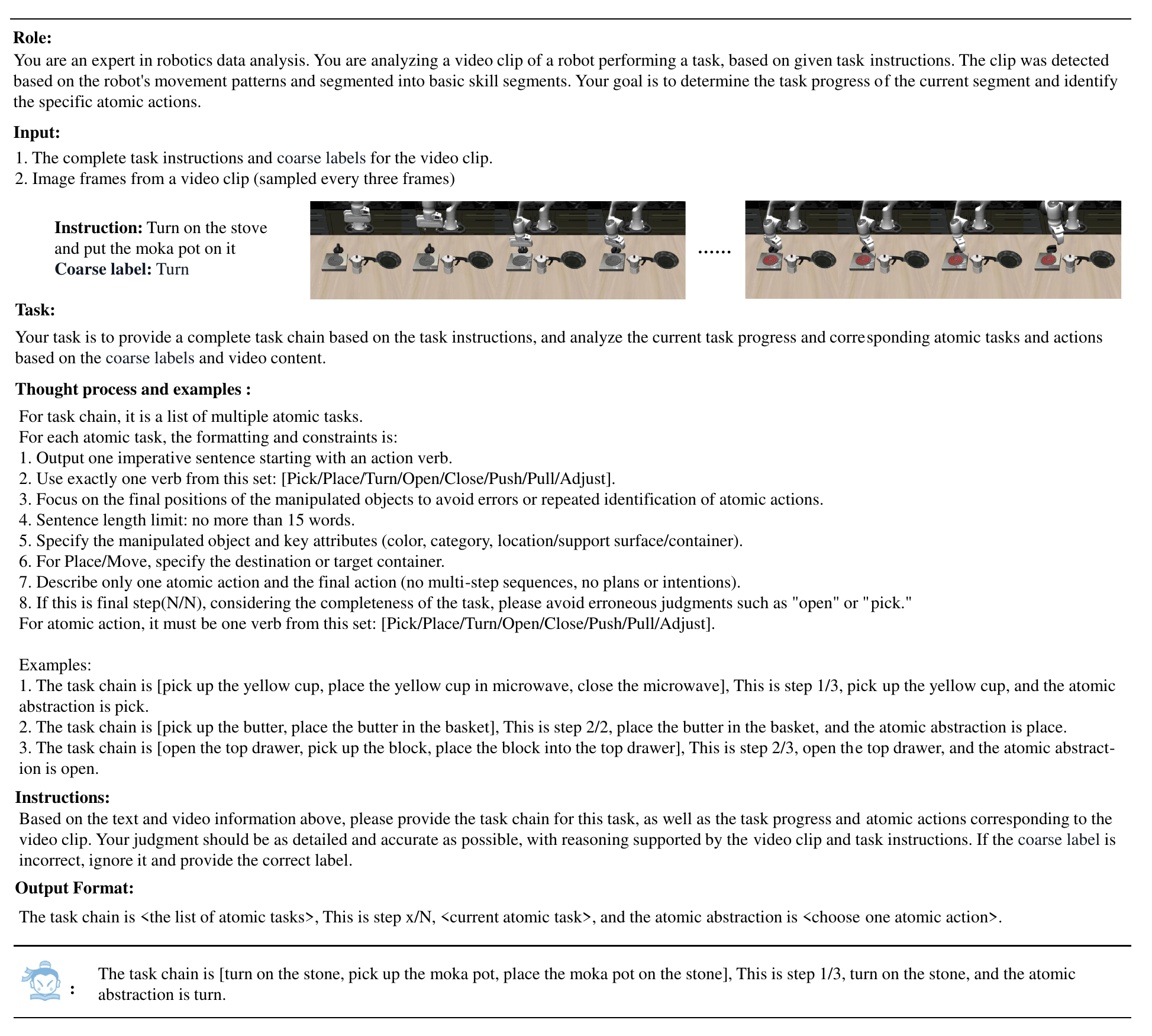} 
    \caption{\tb{The prompts and examples of the InternVideo2.5.}}
    \label{example——intrnvideo} 
    \vspace{-5pt} 
\end{figure*}

\subsection{Simulations Setting}

\mypara{LIBERO Setting.}
We use the public dataset provided by LIBERO and convert it into the Lerobot format for all experiments. Following the data processing method introduced in Sec.~3.5, we perform fine grained annotation and organize the collected data into five atomic action abstractions: $Pick$, $Place$, $Open$, $Close$, and $Turn$. The data distribution for these action categories is presented in Tab.~\ref{tab:libero_atomic_skill_distribution}. All skills are trained in a mixed manner, and therefore maintaining balanced data becomes essential. To achieve this, we increase the sampling frequency of the less represented actions, specifically $Open$, $Close$, and $Turn$, in order to equalize the data distribution and prevent insufficient training of the corresponding skill experts. For a fair comparison, AtomicVLA is consistent with the evaluation of the baseline method, testing each task 50 times and reporting the average results.

\begin{table*}[t]
\centering
\caption{\tb{List of tasks and prompts used in our real-world experiments.}}
\begin{tabular}{ll}
\toprule
\textbf{Task Type} & \textbf{Task Prompt} \\
\midrule
\multicolumn{2}{l}{\textbf{Long-horizon Tasks}} \\
Objects in plate & Place all blocks on the table into a green plate. \\
Object into drawer & Open the top drawer and place the block inside. \\
Object into microwave & Place the plate into the microwave and close the door. \\
\midrule
\multicolumn{2}{l}{\textbf{Short Tasks}} \\
Grasp & Grasp the block from the table. \\ 
Stack & Stack the red block on the orange block. \\
Close & Close the microwave on the table. \\
Press & Press the button on the table. \\
Open & Open the top drawer. \\
\midrule
\multicolumn{2}{l}{\textbf{Complex Scenes}} \\
Objects in plate & Put the pepper and corn into the green plate. \\
Objects in plate & Put the carrot and cucumber into the green plate. \\
Objects in plate & Put the potato and eggplant into the green plate. \\
\bottomrule
\end{tabular}
\label{tab:task_prompts}
\end{table*}

\begin{table*}[t]
\centering
\caption{\tb{Results on Complex Scenes.}}
\label{tab:complex}
\vspace{-3pt}
\addtolength{\tabcolsep}{0.5pt} 
\begin{tabularx}{0.7\linewidth}{l c c c c}
\toprule
\tb{Method}   & \tb{Pepper/Corn} &  \tb{Carrot/Cucumber} & \tb{Potatoe/Eggplant} & \tb{Avg.}  \\
\midrule 
$\pi_{0.5}$~\cite{intelligence2025pi05visionlanguageactionmodelopenworld}        & 25 & 40 & 35 & 33.3     \\
\midrule 
\rowcolor{mygray} \textbf{\name{}*}    & \tb{40} & \tb{45} & \tb{45} & \tb{43.3}  \\
\bottomrule
\end{tabularx}
\end{table*}

\begin{table*}[t]
\centering
\small
\caption{\tb{Success rates for all evaluated tasks on CALVIN ABC-D dataset.}}
\begin{tabular}{l c l c l c}
\hline
\textbf{Task Name} & \textbf{SR (\%)} & \textbf{Task Name} & \textbf{SR (\%)} & \textbf{Task Name} & \textbf{SR (\%)} \\
\hline
rotate blue block right   & 97.4 & lift red block table      & 99.4 & lift blue block table     & 99.4 \\
move slider right         & 100.0 & lift pink block table     & 94.5 & place in drawer           & 100.0 \\
lift red block slider     & 99.3 & move slider left          & 100.0 & rotate red block left     & 98.5 \\
place in slider           & 98.6 & turn on lightbulb         & 100.0 & push pink block left      & 93.5 \\
turn off lightbulb        & 100.0 & rotate blue block left    & 100.0 & lift blue block slider    & 95.6 \\
turn off led              & 98.8 & push blue block left      & 94.2 & lift pink block drawer    & 100.0 \\
push into drawer          & 86.0 & turn on led               & 100.0 & rotate pink block right   & 98.6 \\
lift blue block drawer    & 100.0 & stack block               & 98.4 & unstack block             & 98.6 \\
close drawer              & 100.0 & push pink block right     & 33.8 & push blue block right     & 22.2 \\
lift pink block slider    & 97.8 & push red block right      & 29.2 & rotate pink block left    & 100.0 \\
open drawer               & 100.0 & push red block left       & 89.9 & lift red block drawer     & 100.0 \\
rotate red block right    & 97.3 &                           &       &                           &      \\
\hline
\end{tabular}
\label{tab:calvin_detail}
\end{table*}

\begin{table}[h]
\centering
\caption{\tb{Parameter and inference-time.}}
\vspace{-8pt}
\addtolength{\tabcolsep}{3.5pt}
\begin{tabularx}{0.98\linewidth}{l|cccc}
\toprule
\tb{Experts} & \tb{$\pi_0$} & \tb{K=5} & \tb{K=8} & \tb{K=12} \\
\midrule
Params  & 3.24B & 4.17B & 4.81B & 5.65B \\
\midrule
Act  &  71ms & 92\,ms & 126\,ms & 160\,ms  \\
Think & - & 104\,ms & 104\,ms & 104\,ms \\
\bottomrule
\end{tabularx}
\vspace{-15pt}
\label{B}
\end{table}

\begin{figure*}[t]
    \centering
    \includegraphics[width=0.8\textwidth]{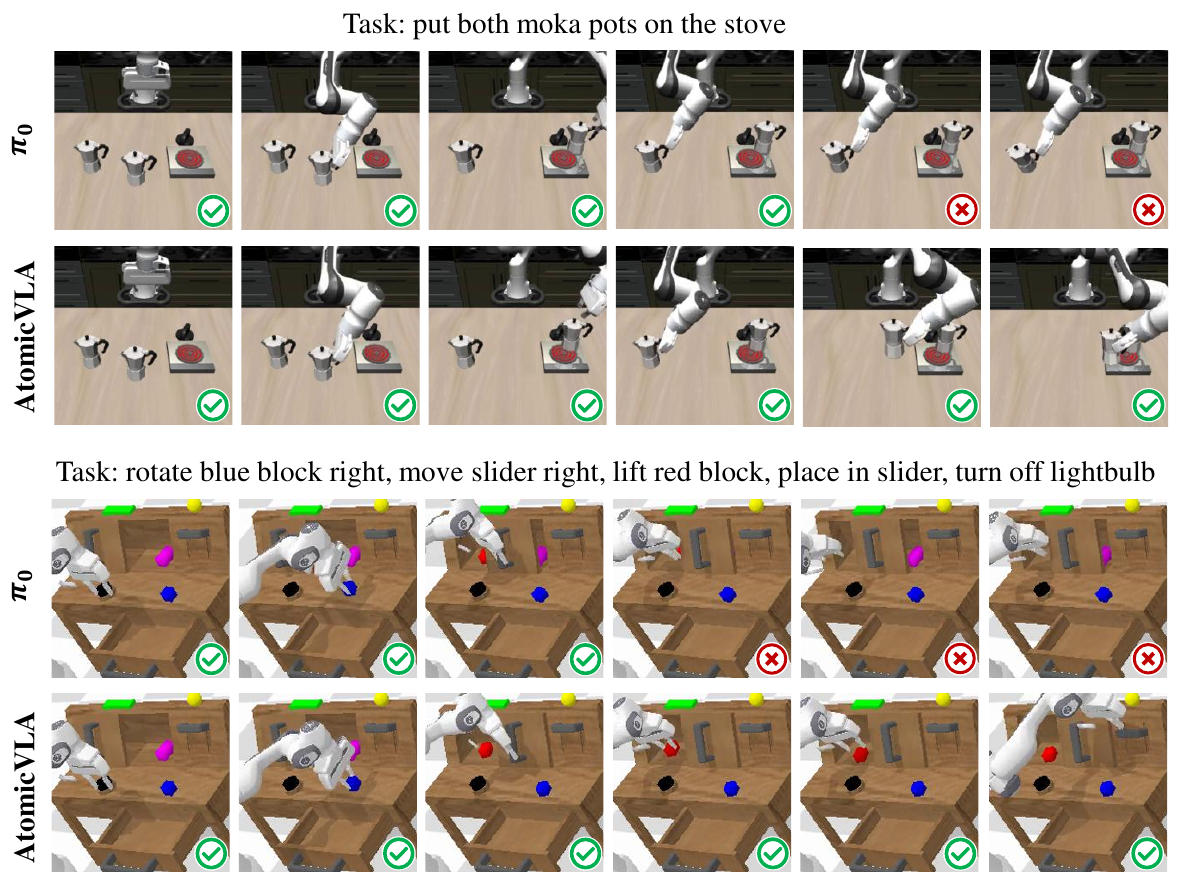} 
    \caption{\tb{Demonstrations of LIBERO and Calvin experiments.}}
    \label{libero&calvin} 
\end{figure*}

\begin{figure*}[t]
    \centering
    \includegraphics[width=0.92\textwidth]{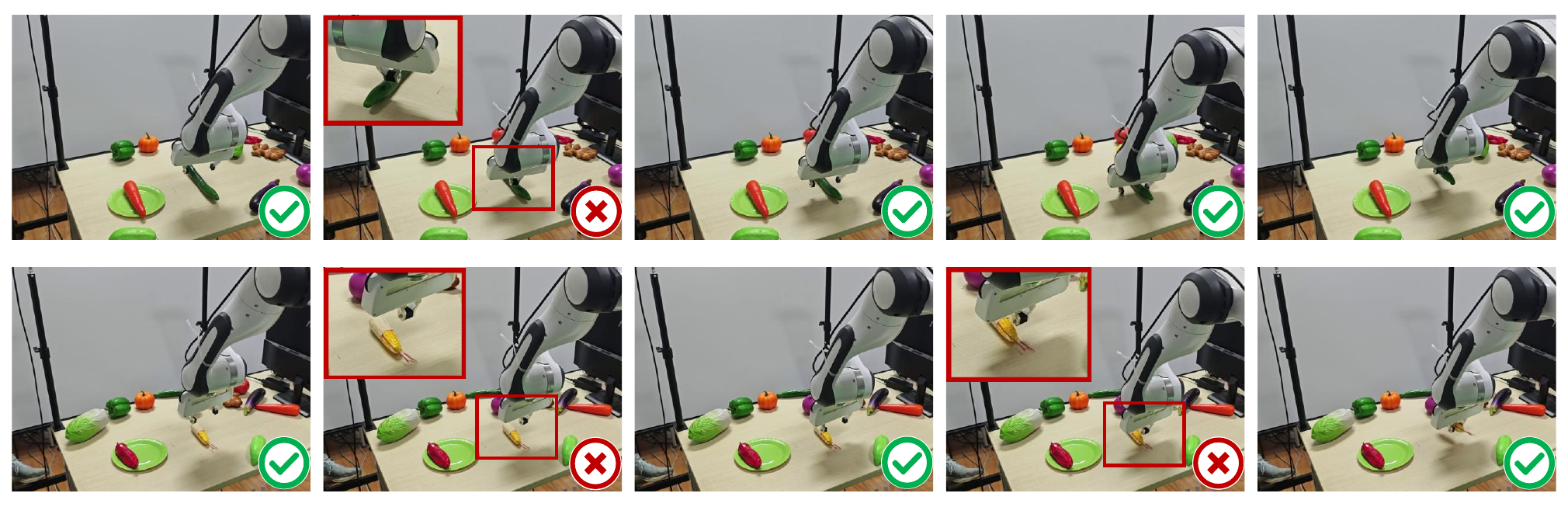} 
    \caption{\tb{Error recovery cases in real-world experiments.}}
    \label{error-real} 
\end{figure*}

\mypara{Calvin Setting.}
We use the task ABC-D public dataset provided by Calvin and divide the data according to the instruction annotations and the corresponding frame intervals. Each trajectory is capped at 64 frames and is converted into the Lerobot format for our experiments. Following the data processing method introduced in Sec.3.5, we perform fine grained annotation and organize the data into eight atomic action abstractions: $Rotate$, $Push$, $Move$, $Open\&Close$, $Lift$, $Place$, $Turn$, and $Stack$. Based on these categories, we construct a skill expert library consisting of 8 skill experts. Building on this configuration, we ensure fair comparison by keeping the AtomicVLA evaluation protocol consistent with that of the baseline methods. In this setting, the robot executes 1,000 task sequences, and each sequence contains five consecutive tasks. We report the average success rates together with the average length of the completed sequences.

\subsection{Real-world Setting}

\mypara{Hardware.}
Our real-world experimental setup consists of a Franka Research3 robotic arm with two Realsense D435i cameras: one mounted on the wrist to provide a first-person perspective, and the other positioned opposite the robotic arm to offer a third-person view. 

\mypara{Evaluation Tasks.}
In the real world, we collected three long-horizon tasks and five short tasks, and additionally gathered three long-horizon tasks in more complex scenarios to evaluate the performance of AtomicVLA. We employed Gello~\cite{wu2024gellogenerallowcostintuitive} to control the Franka arm and record demonstration data. We collected 100 trajectories per long-horizon task and 50 per short task. The results reported in this paper were obtained using a multi-task mixed training protocol. Each task was evaluated 20 times with randomized object placements, and the average performance across these trials was reported as the final test result. The full list of tasks is presented in Tab.~\ref{tab:task_prompts}.


\subsection{Continual Learning Setting}
We conducted experiments on continual learning for short tasks. Specifically, we used four tasks for mixed training, iterating for 20k steps. Then, we applied “open the top drawer" as a new skill for continual learning, fine-tuning on the weights learned from the four tasks. We used a learning rate of \(5 \times 10^{-6}\) and iterated for 7k steps, and reported the results by averaging over 20 validation runs for each of the five tasks.

\subsection{Data Generation Setting}
We use principal component analysis to obtain precise video segmentation and coarse labels. By analyzing the motion changes across five consecutive frames, we determine the dominant motion axis. Specifically, the threshold for the translation axis(\(\Delta x, \Delta y, \Delta z\)) is set to 3 cm, the threshold for the rotation axis(\(\Delta \text{roll}, \Delta \text{pitch}, \Delta \text{yaw}\)) is set to 0.05 radians, and the gripper change( \(\Delta \text{Grip}\)) threshold is set to 0.1.

In Fig.~\ref{example——intrnvideo}, we provide detailed prompts and examples for VLM (InternVideo2.5~\cite{wang2025internvideo25empoweringvideomllms}). The VLM analyzes video clips and generates task chains, task progress, and atomic actions based on the input text instructions.

\section{Additional Results}


\mypara{Detail Results on Calvin.} 
As shown in Tab.~\ref{tab:calvin_detail}, we report the performance of AtomicVLA* on the 34 evaluation tasks of the Calvin ABC-D dataset. The results indicate that the model achieves success rates close to 100 percent on most tasks. However, performance on several $Push$ $blocks$ $right$ tasks is considerably lower, with average success rates only between 20 and 30 percent. Building on this observation, we find that in the training set the relevant blocks are typically placed near the center of the table. In contrast, during evaluation the blocks are often positioned on the right side of the table. This distribution shift leads the model to push the block in the correct direction while failing to push it far enough to satisfy the success criterion, which results in task failure and prevents the execution of subsequent steps.

\mypara{Results on Complex Scenes.} 
As shown in Tab.~\ref{tab:calvin_detail}, we report the performance of AtomicVLA* and $\pi_{0.5}$ on three additional real-world experiments designed to evaluate its ability to handle complex scenes and grasp irregular objects. AtomicVLA* achieved an average accuracy of 43.3\%, which is 10\% higher than the $\pi_{0.5}$ average. In addition, when picking corn, due to the color being similar to the background of the table, AtomicVLA* was able to make multiple corrections as it approached the target, resulting in a 15\% improvement.

\mypara{Parameter and inference-time.} 
Tab.~\ref{B} shows the parameter counts and inference-time on a single H20 GPU. Even with 12 experts, the inference latency is only 160 ms, which is fully practical for real-world use.

\begin{figure*}[t]
    \centering
    \includegraphics[width=1.0\textwidth]{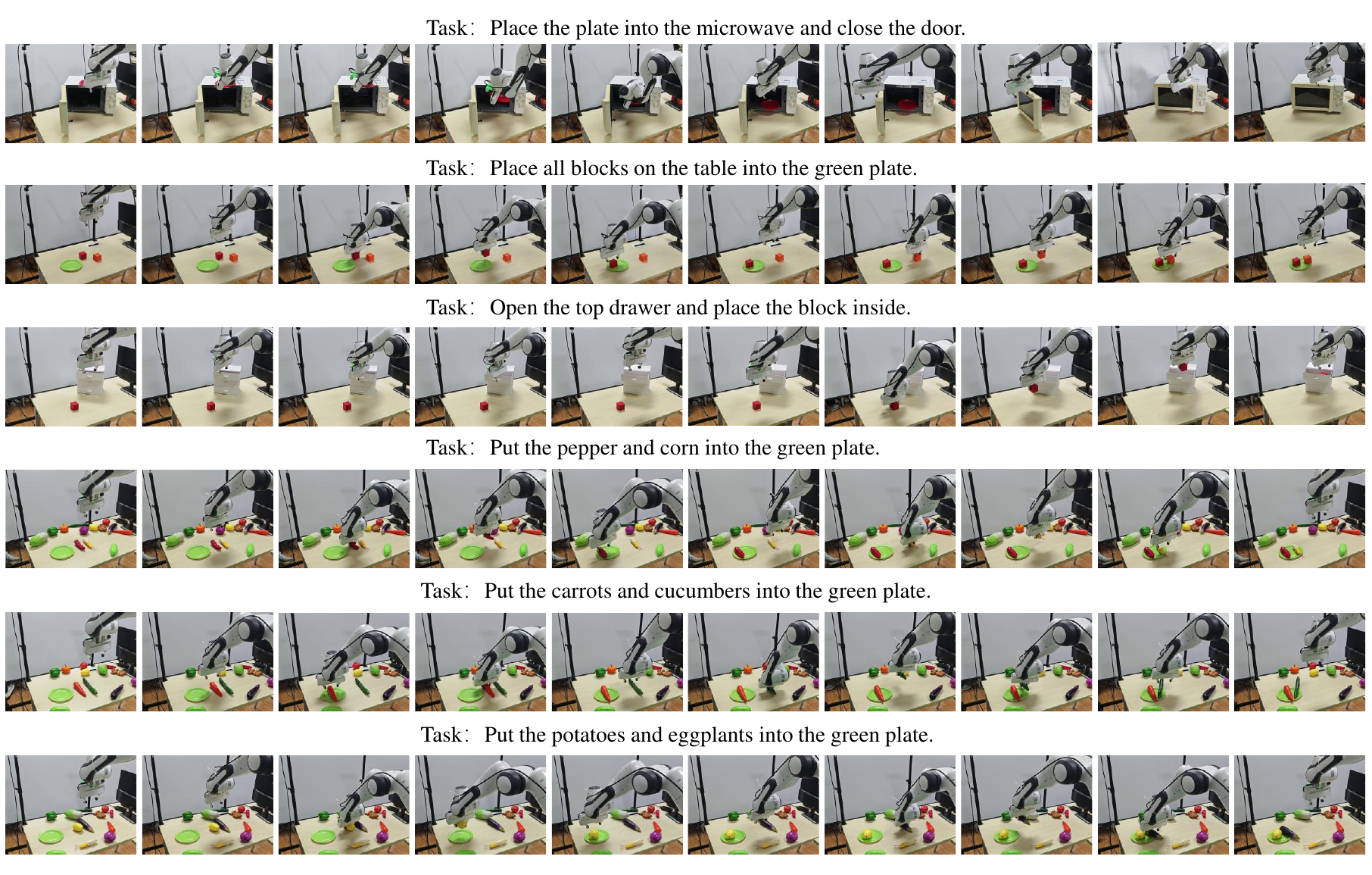} 
    \caption{\tb{Demonstrations of real-world experiments(Long-horizon tasks).}}
    \label{long-vis} 
\end{figure*}

\section{Additional Visualizations}
In Fig.~\ref{libero&calvin}, we present a comparison between AtomicVLA and $\pi_{0}$ across simulation environments. Representative task cases are selected from both LIBERO and Calvin. As shown, AtomicVLA successfully completes several task instances where pi0 fails, demonstrating its stronger robustness and execution reliability in simulated settings.

In Fig.~\ref{error-real}, we further illustrate AtomicVLA’s real-world error recovery capability. When a subtask fails during execution, AtomicVLA automatically replans and corrects its behavior to ensure successful completion of the overall task. Specifically, as highlighted in the red box in the figure, when execution errors occur, such as misgrasps due to inaccurate positioning or visual ambiguity between the target object and the background, AtomicVLA can assess the current task state, generate an updated task plan, and reattempt the failed subtask, thereby ensuring robust completion of the overall task. 

Additionally, we show more demonstrations of real-world experiments in Fig.~\ref{long-vis}. These experiments span a wide spectrum of scenarios, from simple to highly complex tasks and from regular to irregular objects. Across all settings, AtomicVLA consistently exhibits strong performance and robust generalization.

\end{document}